\documentclass{article}

\usepackage[preprint]{neurips_2024}

\usepackage[utf8]{inputenc} 
\usepackage[T1]{fontenc}    
\usepackage{hyperref}       
\usepackage{url}            
\usepackage{booktabs}       
\usepackage{amsfonts}       
\usepackage{nicefrac}       
\usepackage{microtype}      
\usepackage{xcolor}         
\usepackage{graphicx}
\usepackage{subfig}
\usepackage{hyperref}
\usepackage{amsmath}
\usepackage{amssymb}
\usepackage{amsfonts}
\usepackage{tabularx}
\usepackage{comment}

\newcommand*{\affmark}[1][*]{\textsuperscript{$#1$}}

\title{SMAC-Hard: Enabling Mixed Opponent Strategy Script and Self-play on SMAC}

\author{
Yue Deng$^{1}$,
Yan Yu$^{2}$,
Weiyu Ma$^{3}$,
Zirui Wang $^{1}$,
Wenhui Zhu $^{4}$,
Jian Zhao $^{4}$,
Yin Zhang $^{1}$\\
\affmark[1]College of Computer Science and Technology, Zhejiang University \\
\affmark[2]University of Science and Technology of China\\
\affmark[3]Institute of Automation, Chinese Academy of Sciences\\
\affmark[4]Polixir\\
\texttt{devindeng@zju.edu.cn, yy1140730050@mail.ustc.edu.cn, maweiyu2022@ia.ac.cn}\\
\texttt{ziseoiwong@zju.edu.cn, wenhui.zhu@polixir.ai, jian.zhao@polixir.ai}\\
\texttt{zhangyin98@zju.edu.cn}
}

\begin{document}

\maketitle

\begin{abstract}

The availability of challenging simulation environments is pivotal for advancing the field of Multi-Agent Reinforcement Learning (MARL). In cooperative MARL settings, the StarCraft Multi-Agent Challenge (SMAC) has gained prominence as a benchmark for algorithms following centralized training with decentralized execution paradigm. However, with continual advancements in SMAC, many algorithms now exhibit near-optimal performance, complicating the evaluation of their true effectiveness. To alleviate this problem, in this work, we highlight a critical issue: the default opponent policy in these environments lacks sufficient diversity, leading MARL algorithms to overfit and exploit unintended vulnerabilities rather than learning robust strategies. To overcome these limitations, we propose \textbf{SMAC-HARD}, a novel benchmark designed to enhance training robustness and evaluation comprehensiveness. SMAC-HARD supports customizable opponent strategies, randomization of adversarial policies, and interfaces for MARL self-play, enabling agents to generalize to varying opponent behaviors and improve model stability. Furthermore, we introduce a black-box testing framework wherein agents are trained without exposure to the edited opponent scripts but are tested against these scripts to evaluate the policy coverage and adaptability of MARL algorithms. We conduct extensive evaluations of widely used and state-of-the-art algorithms on SMAC-HARD, revealing the substantial challenges posed by edited and mixed strategy opponents. Additionally, the black-box strategy tests illustrate the difficulty of transferring learned policies to unseen adversaries. We envision SMAC-HARD as a critical step toward benchmarking the next generation of MARL algorithms, fostering progress in self-play methods for multi-agent systems. Our code is available at \href{https://github.com/devindeng94/smac-hard}{https://github.com/devindeng94/smac-hard}.

\end{abstract}

\section{Introduction}

Recent advances in Multi-agent reinforcement learning (MARL) have led to significant progress in a wide range of applications such as autonomous vehicle teams \citep{cao2012overview} and sensor networks \citep{zhang2011coordinated}. The domain of MARL has experienced rapid progress, significantly driven by the advent of advanced simulation environments that serve as benchmarks to assess the effectiveness of novel algorithms. These benchmarks play a pivotal role in bridging the gap between theoretical developments and practical implementations by enabling the evaluation of MARL strategies in scenarios that reflect real-world complexities. One of the most influential platforms in this space is the StarCraft Multi-Agent Challenge (SMAC) \citep{samvelyan2019starcraft}, which provides a suite of micromanagement tasks set within the StarCraft II environment.

While SMAC initially gained widespread adoption as a benchmark, its limitations have become apparent with the advancement of MARL techniques \citep{gorsane2022towards}. Many algorithms now attain near-optimal performance on SMAC's pre-defined tasks \citep{hu2021rethinking,yu2022surprising}, raising concerns about its ability to distinguish genuinely innovative approaches from those that exploit specific task structures. This phenomenon of benchmark saturation highlights the need for more intricate and diverse challenges to evaluate the robustness and generalizability of MARL algorithms effectively.

To address these limitations, SMACv2 \citep{ellis2024smacv2} was introduced, offering increased complexity, scalability, and realism through procedurally generated scenarios. These scenarios require agents to generalize to previously unseen conditions during testing, thereby discouraging overfitting and encouraging the development of more adaptable algorithms. However, the added complexity of SMACv2 has also introduced new difficulties, such as challenges in achieving convergence for many MARL algorithms, suggesting that while it presents a tougher benchmark, it may currently exceed the capabilities of existing methodologies \citep{singh2023much}.

Another critical issue with current benchmarks lies in the limited diversity of default opponent policies. This constraint often results in overfitting, where MARL algorithms exploit specific weaknesses in static opponents rather than learning robust strategies \citep{mitra2024towards}. Consequently, these benchmarks may fall short of preparing algorithms for the variability and unpredictability encountered in real-world applications. To overcome this, there is a pressing need for benchmarks that feature a broader spectrum of opponent behaviors, promoting the creation of more resilient and adaptable algorithms.

Motivated by these considerations, we propose SMAC-HARD, an innovative benchmark tailored to enhance the evaluation of MARL algorithms. SMAC-HARD introduces features such as opponent strategy editing, random selection of opponent strategies, and self-play interfaces, requiring agents to generalize across diverse opponent behaviors during training. Furthermore, it includes a black-box testing framework that evaluates policy coverage by testing agents against previously unseen opponent scripts. These enhancements provide a robust evaluation pipeline, facilitating the development of MARL algorithms equipped to handle diverse and unpredictable environments with greater stability and reliability.

\section{Related Work}

\subsection{MARL Environments}
The development of MARL is inseparable from a variety of multi-agent environments.
\citep{Peng_Rashid_Witt_Kamienny_Torr_Böhmer_Whiteson_2020} adapted multi-agent systems to the MuJoCo environment by decomposing the robot's joints and controlling them collaboratively.
MPE \citep{Lowe_Wu_Tamar_Harb_Abbeel_Mordatch_2017} involves controlling the movement of different particles in a 2D space to complete a series of tasks, which presents simple communication in the cooperation and competition among agents.
PettingZoo \citep{Terry_Black_Grammel_Jayakumar_Hari_Sullivan_Santos_Perez_Horsch_Dieffendahl_et_al} is a library of MARL environments, encompassing a variety of game types that allow for competitive, cooperative and mixed agent relations. 
It includes classic scenarios such as Atari \citep{Mnih_Kavukcuoglu_Silver_Graves_Antonoglou_Wierstra_Riedmiller_2013}, board games, and particle control.
In addition, many classic human games have also been applied in MARL research.
Overcooked-AI \citep{Micah_Shah_Ho_Griffiths_Seshia_Abbeel_Dragan_2019}, based on the popular human game, expects agents to learn task distribution and coordination to achieve high rewards in a cooking scenario.
Google Research Football \citep{Kurach_Raichuk_Stańczyk_Zając_Bachem_Espeholt_Riquelme_Vincent_Michalski_Bousquet_et} offers a physics-based 3D soccer simulation, which presents a challenging RL problem as soccer requires balancing short-term control and skill learning (such as passing) with higher-level strategies.
HOK \citep{Wei_Chen_Ji_Qin_Deng_Li_Wang_Zhang_Yu_Liu_et} is a high-complexity Multiplayer Online Battle Arena (MOBA) environment. Players compete by gathering resources while interfering with their opponents to win the game. With multiple heroes and complex state and action spaces, HOK provides an excellent environment for academic research on complex control problems.
SMAC \citep{Samvelyan_Rashid_Witt_Farquhar_Nardelli_Rudner_Hung_Torr_Foerster_Whiteson_2019} is one of the most popular environments for MARL. Based on the widely known real-time strategy (RTS) game StarCraft II, SMAC scenarios are carefully designed and require learning one or more micromanagement techniques to defeat enemies. Each scenario involves a confrontation between two armies, with varying initial positions, unit types, and terrain features like high ground or impassable obstacles.
SMACv2 \citep{Ellis_Moalla_Samvelyan_Sun_Mahajan_Foerster_Whiteson} addresses the issue of randomness in SMAC by generating unit types for each agent with a fixed probability distribution, making the environment more challenging. This significantly enhances the diversity and complexity of the environment, providing a richer setting for contemporary 
\subsection{Algorithms for MARL}
Value-based MARL algorithms have made significant progress in recent years. IQL \citep{Tampuu_Matiisen_Kodelja_Kuzovkin_Korjus_Aru_Aru_Vicente_2017} treats other agents as part of the environment and trains independent Q-value networks for agents. 
However, it may not always converge for the non-stationarity of the environment caused by the changing policies of other agents. 
A series of methods (VDN \citep{Sunehag_Lever_Gruslys_Czarnecki_Zambaldi_Jaderberg_Lanctot_Sonnerat_Leibo_Tuyls_et}, QMIX \citep{Rashid_Samvelyan_Witt_Farquhar_Foerster_Whiteson_2018}, QTRAN \citep{Son_Kim_Kang_Hostallero_Ye_2019}, QPLEX\citep{Wang_Ren_Liu_Yu_Zhang_2020}) decompose the global Q-value function into individual Q-value functions to each agent.
LDSA \citep{Yang_Zhao_Hu_Zhou_Li_2022} proposes an ability-based subtask selection strategy to assign agents to different subtasks reasonably and dynamically group agents with similar abilities into the same subtask.
Some other algorithms apply TRPO \citep{Schulman_Levine_Moritz_Jordan_Abbeel_2015} or PPO \citep{Schulman_Wolski_Dhariwal_Radford_Klimov_2017} to multi-agent problems.
For instance, IPPO \citep{Witt_Gupta_Denys_Makoviychuk_Torr_Sun_Whiteson_2020} apply PPO and enforce parameter sharing under the assumption that all agents have the same action space.
MAPPO \citep{Yu_Velu_Vinitsky_Wang_Bayen_Wu_2021} enhances IPPO by introducing a joint critic function and improving the implementation techniques. 
HATRPO and HAPPO \citep{Kuba_Chen_Wen_Wen_Sun_Wang_Yang_2021} extend the theory of trust region learning to cooperative MARL, which holds in general and does not require any assumption that agents share parameters or the joint value function is decomposable.
\subsection{LLM for Decision Making}
The emergence of large language models (LLMs) has significantly advanced the research on decision-making for agents in complex environments. LLMs leverage massive human data to gain a deep understanding of various complex scenarios, including robotic control, gaming, etc. 
In robotics, \citep{Liang_Huang_Xia_Xu_Hausman_Ichter_Florence_Zeng_2022} proposed "code as policy" for robotic control. In this approach, LLMs generate Python code to process sensory outputs and parameterize control primitives. 
The Eureka algorithm \citep{Ma_Liang_Wang_Huang_Bastani_Jayaraman_Zhu_Fan_Anandkumar_2023} utilizes LLMs for human-level reward design in reinforcement learning tasks, which ables to generate reward functions that surpass those designed by experts. 
LLMs have also shown exceptional performance across a wide range of game types, from classic board games to open-world video games.
ChessGPT \citep{Feng_Luo_Wang_Tang_Yang_Shao_Mguni_Du_Wang_2023} demonstrated the ability of LLM agents to understand and strategize in strategic games. 
\citep{jin2024learning} and \citep{Xu_Yu_Fang_Wang_Wu_2023} has explored the social deduction game such as Werewolf, which presented unique challenges in strategic communication and decision-making.
The MineDojo environment \citep{Fan_Wang_Jiang_Mandlekar_Yang_Zhu_Tang_Huang_Zhu_Anandkumar_2022} facilitated projects like GITM \citep{Zhu_Chen_Tian_Tao_Su_Yang_Huang_Li_Lu_Wang_et} and Voyager \citep{Wang_Xie_Jiang_Mandlekar_Xiao_Zhu_Fan_Anandkumar_Nvidia_Caltech_et} in Minecraft, demonstrating LLM agents' ability to navigate and perform tasks in complex 3D environments.
The Cradle framework \citep{Tan_Zhang_Xu_Xia_Ding_Li_Zhou_Yue_Jiang_Li_et} introduced a novel approach allowing LLM agents to interact with various software and games through a unified interface of screenshots and keyboard/mouse inputs. This demonstrated the potential for general-purpose game-playing agents across multiple commercial video games and software applications.
\section{Background}

\subsection{MARL}

A fully cooperative multi-agent task is described as a Dec-POMDP \citep{Oliehoek_Amato_2016} task which consists of a tuple $G = \langle S, A, P, r, Z, O, N, \gamma\rangle$ in which $s\in S$ is the true state of the environment in the centralized training phase and $N$ is the number of agents. 
At each time step, each agent $i \in N \equiv \{1,\dots,n\}$ chooses an action $a_i \in A$ which forms the joint action $\mathbf{a} \in \mathbf{A} \equiv A^N$. 
The transition on the environment is according to the state transition function that $P(\cdot|s, \mathbf{a}):S\times \mathbf{A} \times S \to [0, 1]$. 
The reward function, $r(s, \mathbf{a}):S \times A \to \mathbb{R}$, is shared among all the agents, and $\gamma \in [0,1)$ is the discount factor for future reward penalty. 
Partially observable scenarios are considered in this paper that each agent draws individual observations $z \in Z$ of the environment during the decentralized execution phase according to the observation functions $O(s,i):S \times N \to Z$. 
Meanwhile, the action-observation history, $\tau_i \in T \equiv (Z \times A)^*$, is preserved for each agent and conditions the stochastic policy $\pi_i(a_i|\tau_i):T \times A \to [0,1]$. 

Value-based MARL algorithm aims to find the optimal joint action-value function $Q^*(s,\mathbf{a};\theta)=r(s,\mathbf{a})+\gamma \mathbb{E}_{s'}\left[\max_{\mathbf{a'}} Q^*\left(s',\mathbf{a'};\theta\right)\right]$ and parameters $\mathbf{\theta}$ are learned by minimizing the expected TD error. 
VDN learns a joint action-value function $Q_{tot}(\mathbf{\tau},\mathbf{a})$ as the sum of individual value functions: $Q^{\text{VDN}}_{tot}(\mathbf{\tau}, \mathbf{a}) = \sum_{i=1}^n Q_i(\tau_i, a_i)$. 
QMIX introduces a monotonic restriction $\forall i \in N, \frac{\partial Q^{\text{QMIX}}_{tot}(\mathbf{\tau}, \mathbf{a})}{\partial Q_i(\tau_i, a_i)} > 0$ to the mixing network to meet the IGM assumption.
In policy-based algorithms, agents use a policy $\pi_\theta(a_i|\tau_i)$ parameterized by $\theta$ to produce an action $a_i$ from the local observation and jointly optimize the discounted accumulated reward $J(\theta)=\mathbb{E}_{a^t,s^t}[\sum_t\gamma^tr(s^t,a^t)]$ where $a^t$ is the joint action at time step $t$.
In the AC-based algorithm, MAPPO algorithm, the actor is updated by optimizing the target function $J_{\theta^k}(\theta)=\sum_{s^t,a^t}\min(\frac{\pi_\theta(a^t|s^t)}{\pi_{\theta^k}(a^t|s^t)}A_{\theta^k}(s^t,a^t), clip(\frac{\pi_\theta(a^t|s^t)}{\pi_{\theta^k}(a^t|s^t)},1-\epsilon,1+\epsilon)A_{\theta^k}(s^t,a^t))$, where the $\epsilon$ is the clip parameter and $A_{\theta^k}(s^t,a^t)$ is the advantage function.
The critic training is similar to value-based $Q$ learning by calculating TD-error and TD targets.
During the TD training process, the target value is calculated by bootstrapping from the existing Q-function according to temporal difference methods or Monte Carlo returns.

\subsection{SMAC}

Instead of addressing the complexities of the full StarCraft II game, the StarCraft Multi-Agent Challenge (SMAC) concentrates on micromanagement scenarios where each military unit is controlled by an individual learning agent. During testing, units are restricted by a limited field-of-view and lack explicit communication mechanisms. Featuring a diverse range of challenging setups, SMAC has become a widely adopted benchmark in the MARL community for evaluating algorithms. It includes 23 micromanagement scenarios categorized into three types: symmetric, asymmetric, and micro-trick. Symmetric scenarios involve equal numbers and types of units for both allies and enemies. Asymmetric scenarios present additional units for the enemy, making them more challenging. Micro-trick scenarios require specialized strategies, such as in 3s\_vs\_5z, where three allied stalkers must kite five zealots, or in corridor, where six zealots must block a narrow passage to defeat 24 zerglings without being overwhelmed.

SMACv2 builds upon the original SMAC by addressing its shortcomings and providing a more robust framework for cooperative MARL evaluation. It emphasizes diversity, scalability, and realism in multi-agent tasks while resolving the exploitable weaknesses of its predecessor. SMACv2 introduces new maps with varied unit compositions, asymmetrical team setups, and heterogeneous agents, offering more intricate and balanced challenges. Additionally, it incorporates stochastic elements in agent behaviors and environmental dynamics, fostering the development of algorithms capable of handling uncertainty. Empirical evaluations highlight SMACv2 as a significantly more demanding benchmark, encouraging research into generalizable MARL approaches. By setting a higher standard, SMACv2 promotes innovation and ensures advancements are relevant to real-world multi-agent systems.

\section{Limitation on Default Opponent Strategy}

In this section, we examine how the variation in opponents' strategies affects the performance of two widely used baseline algorithms, QMIX and MAPPO. Our analysis demonstrates that relying on a single deterministic opponent strategy can cause the MARL training process to overfit to a particular policy, leading to a reduction in the generalization capabilities of the trained models. Additionally, when the opponent employs a single deterministic strategy that has certain weaknesses, agents may exploit these vulnerabilities, resulting in suboptimal solutions.

\subsection{Default Opponent Policy}

\begin{figure}[h!]
    \centering
    \subfloat[]{\includegraphics[width=0.49\columnwidth]{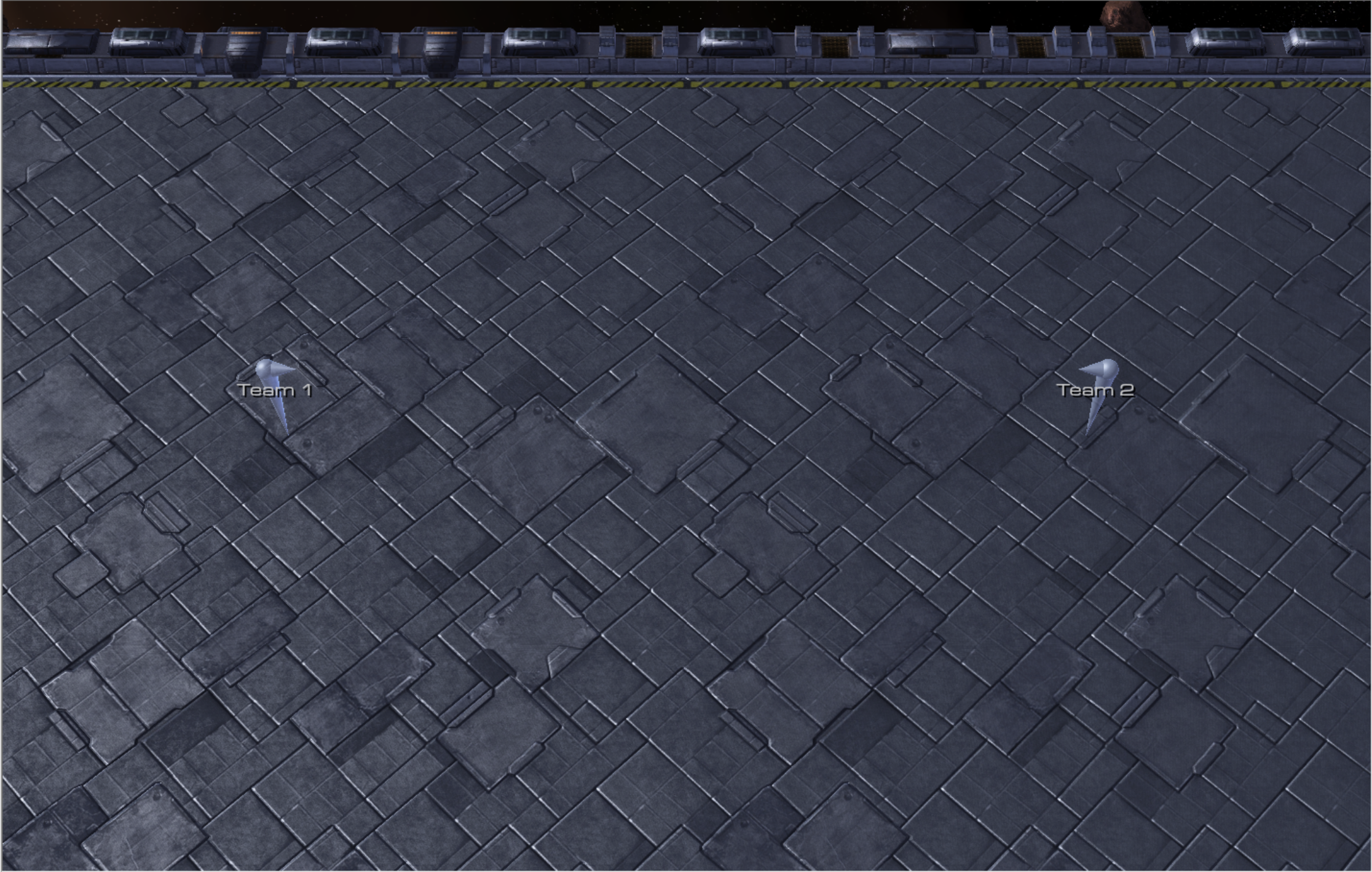}} 
    \hspace{2px}
    \subfloat[]{\includegraphics[width=0.49\columnwidth]{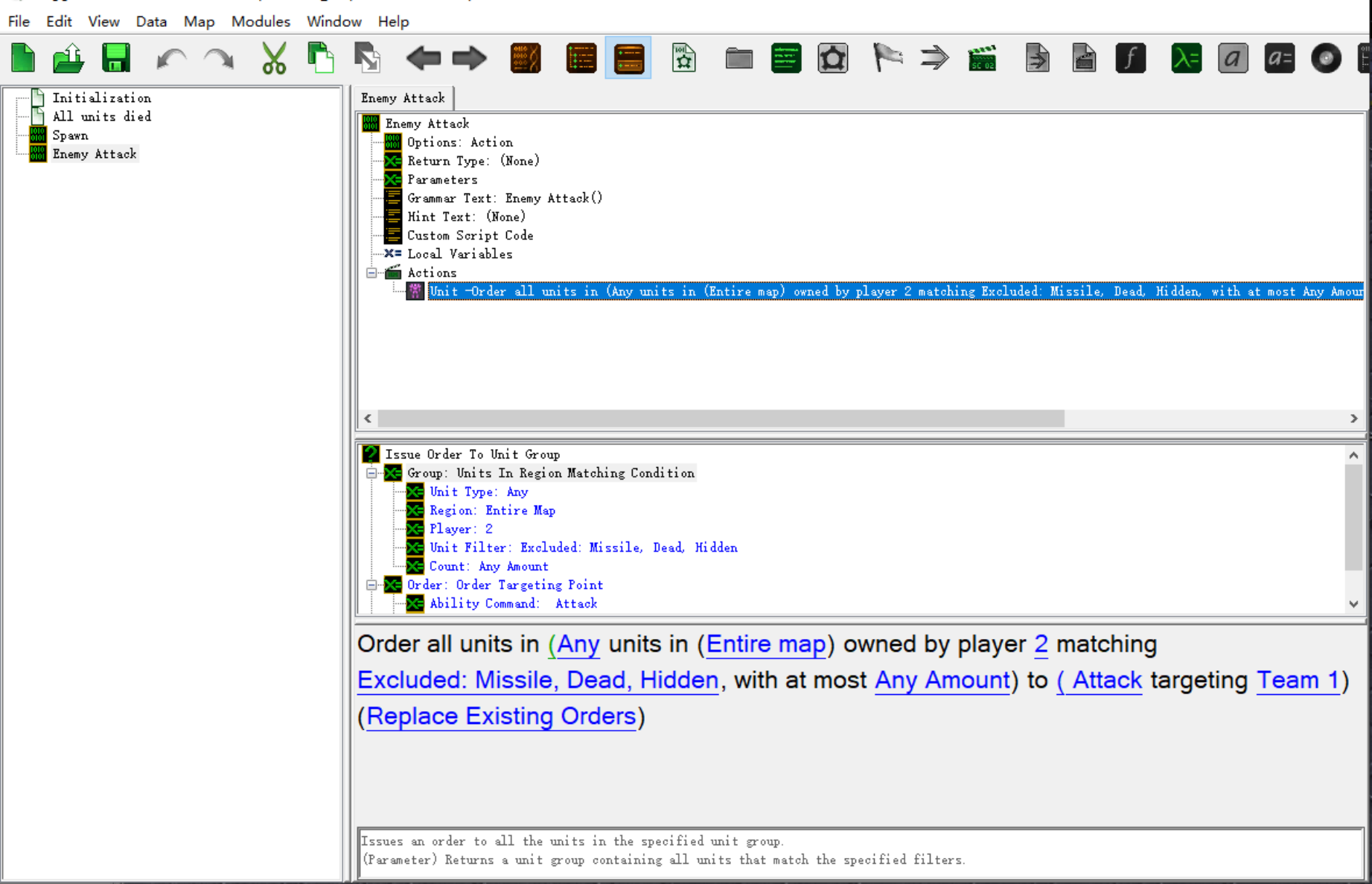}} 
    \caption{(a) Agents spawn at the Team 1 point and the opponent units spawn at Team 2 point. (b) The internal opponent script is defined in the SC2Map file. The opponent controls all the units to attack toward the Team 1 point. \textit{Order all units in (Any units in (Entire map) owned by player 2 matching Excluded: Missile, Dead, Hidden, with at most Any Amount) to (Attack targeting Team 1)(Replace Existing Orders)}.}
    \label{fig:single-policy} 
\end{figure}

In the StarCraft II (SC2Map) configuration, the default opponent policy is defined within the map files rather than the Python environment of SMAC. These map files can be edited using the StarCraft II Editor, which is officially provided by Blizzard. As illustrated in Figure \ref{fig:single-policy}, most map configurations specify two spawning positions for both agents and their opponents. An internal script directs the opponent units to move towards the agents' starting positions. The opponent units will automatically target and attack the nearest agents within their sight range, based on a hate value. This causes the opponent units to continuously adjust their targets.

\begin{figure}[h!]
    \centering
    \subfloat[]{\includegraphics[width=0.45\columnwidth]{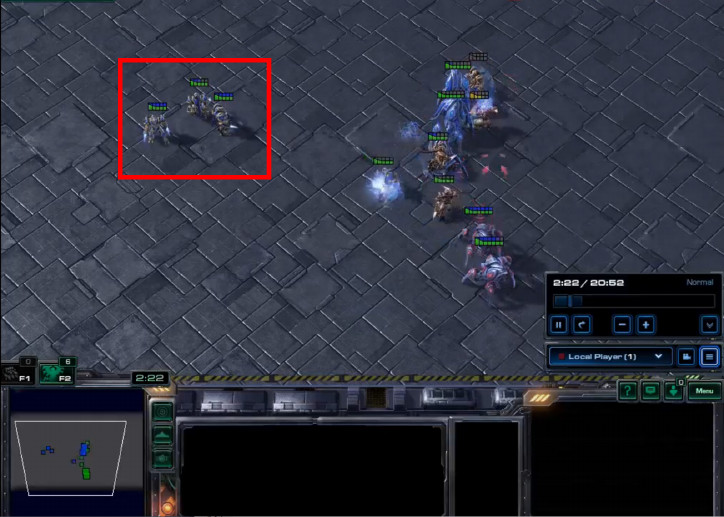}} 
    \subfloat[]{\includegraphics[width=0.45\columnwidth]{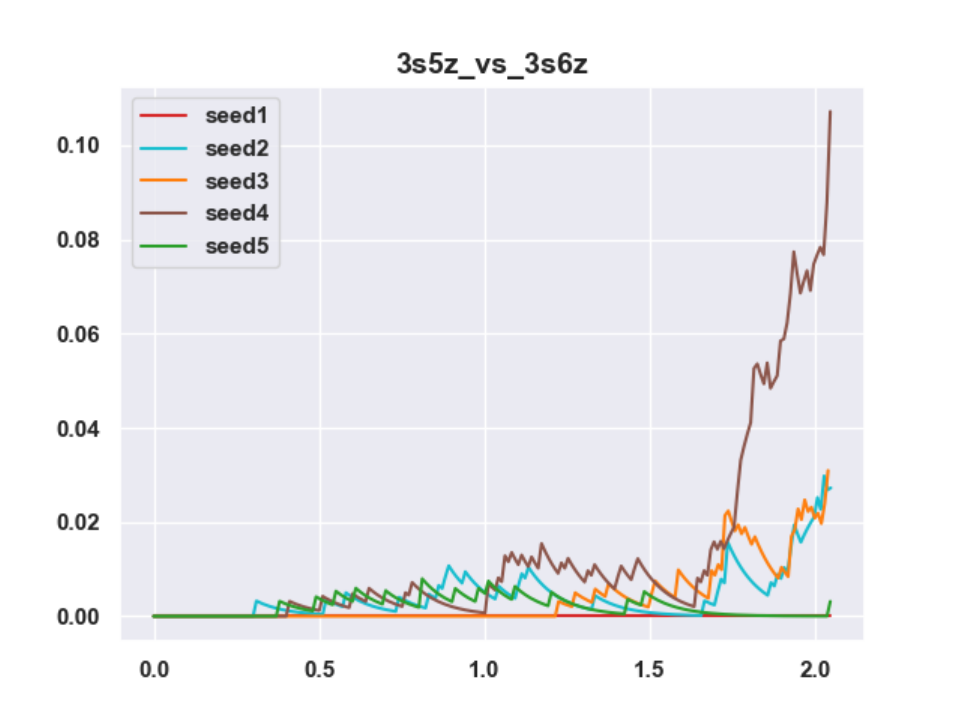}} 
    \caption{(a) Three opponent Zealot units arrive at the agents' starting point and are stuck at that point. (b) MARL algorithms easily achieve high performance when the tricky strategy is explored.}
    \label{fig:bug} 
\end{figure}

Typically, MARL models direct agents toward the opponent’s side and execute micro-management tasks based on observations. However, as depicted in Figure \ref{fig:bug}, there are situations where the opponent agents are not drawn to any hate values and end up becoming stuck at their designated destination points. This flaw in the default opponent behavior presents an opportunity for the agents to exploit a strategy of initially hiding from the opponent and then attacking each opponent unit individually. This is particularly evident in the 3s5z\_vs\_3s6z map, where many MARL models exploit this vulnerability to secure victories. Consequently, the default opponent policy is incomplete in terms of its policy space, rendering it insufficient for adequately evaluating MARL algorithms.

\subsection{Performance on Mixing Policies}

Recent advancements in Reinforcement Learning (RL) have resulted in substantial improvements in tackling complex control systems, including applications in robotics and Atari games. However, one challenge that persists in RL is the tendency for the models to overfit to the specific strategies of the fixed opponent, limiting the transferability of the learned policies. As a result, RL models face difficulties when adapting to new tasks or even when encountering different opponent strategies within the same task.

\begin{figure}[h!]
    \centering
    \includegraphics[width=0.99\textwidth]{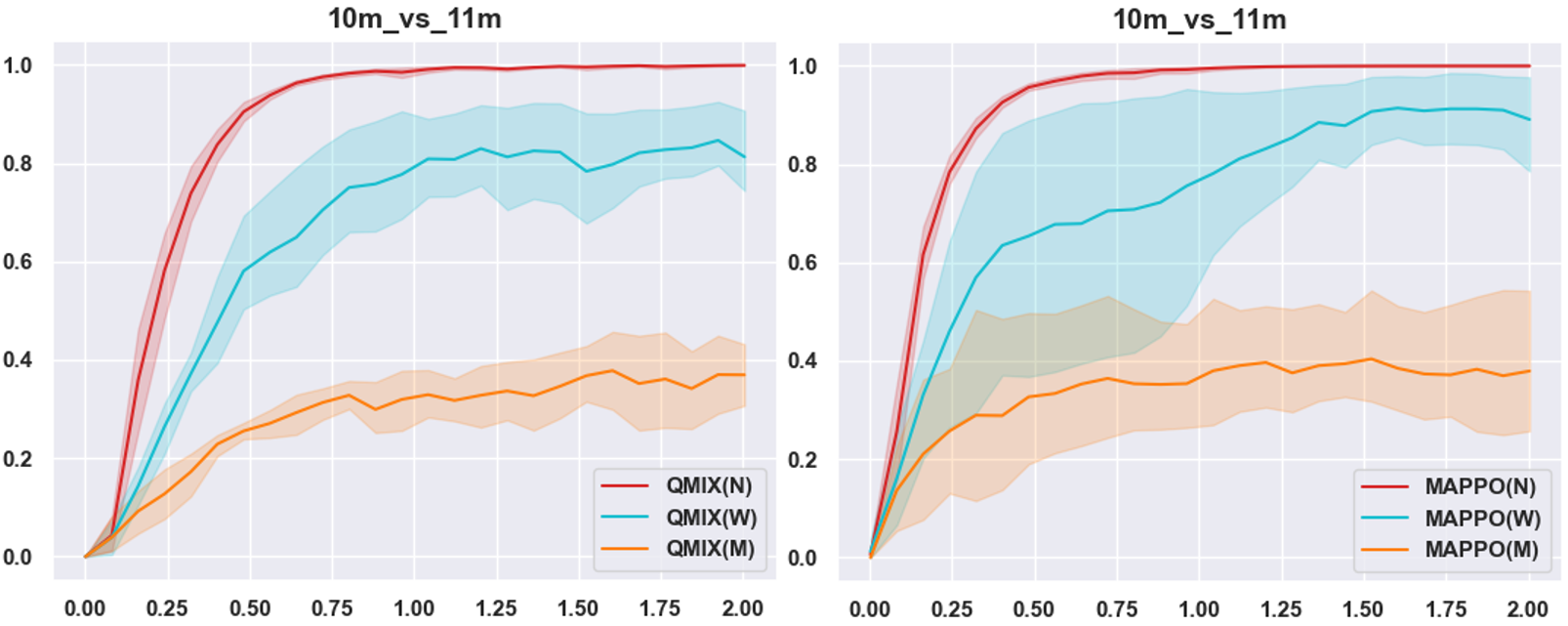}
    \caption{The learning curve of the models from QMIX and MAPPO algorithms when facing 'attacking the nearest enemy' (N), 'attacking the weakest enemy' (W), and the 'randomly choosing from the two strategies' (M). The x-axis is the time steps (1e6) being evaluated and the y-axis is the average winning rate of 5 different seeds from 32 evaluation processes.}
    \label{fig:mixed} 
\end{figure}

To illustrate this phenomenon, we perform an experiment on the 10m\_vs\_11m scenario using two distinct, slightly varied opponent strategies alongside a mixed strategy. We employ the QMIX and MAPPO algorithms, utilizing the default hyperparameter settings outlined in the Appendix, and run the simulation for 2 million time steps. Unlike the original script, where the enemies are drawn towards each other based on hate values, we simplify the setup by modifying the opponent's strategy. The three strategies tested are: 1) focusing fire on the closest enemy, 2) targeting the enemy with the lowest health and shield, and 3) randomly selecting one of the two strategies. The resulting learning curves are presented in Figure \ref{fig:mixed}.

As demonstrated in Figure \ref{fig:mixed}, the MARL algorithms rapidly converge to optimal strategies when facing individual opponent strategies. Both QMIX and MAPPO achieve either perfect or near-perfect performance within 2 million time steps. However, when the algorithms encounter the mixed strategy, their winning rates drop significantly to $36.88\%$ for QMIX and $39.38\%$ for MAPPO. Additionally, the average Q-values at the 2 million time step are $1.03$ for the strategy targeting the nearest enemy, $0.76$ for attacking the weakest enemy, and $0.58$ for the mixed strategy. This indicates that the mixed strategy induces a more conservative approach in the predicted Q-values.

More intensively, the analysis of the replays generated by the three strategies reveals that the learned skills of fire focusing, kiting, and health management are essential response tactics. In scenarios where the opponent employs a strategy of attacking the nearest enemy, an agent with lower health can temporarily retreat to minimize exposure to enemy fire. Conversely, when facing the strategy of targeting the weakest enemy, the weakest agent should actively move forward to draw the opponent's fire, thereby allowing the other agents to carry out their attacks. However, when confronted with a mixed strategy, agents must devote additional time to assessing whether they should continue retreating, based on the likelihood of the opponent’s chosen strategy.

Therefore, training a single MARL model to handle diverse opponent strategies proves to be a significant challenge for current baseline algorithms. To address this, we extend the SMAC environment by introducing SMAC-HARD, which supports mixed opponent scripts to enhance the model's ability to distinguish between different strategies. Additionally, these opponent scripts can be customized through the pysc2 package grammar. To further diversify the opponent strategies, we align the observation, action, and reward interfaces of the opponent agents with those of the learning agents. This alignment facilitates training agents' policies using self-play methods or alternative MARL algorithms, promoting more complex and adaptive learning dynamics.

\section{SMAC-HARD}

The default opponent strategy in SMAC and SMACv2, as previously discussed, is determined by the map configuration, which limits the variety of policies for each task. To overcome these limitations, we introduce three key modifications: an opponent script editing interface, random selection of scripts, and alignment of the opponent MARL interface with the agent's interface. These changes incrementally enhance the diversity of the opponent strategies, making the tasks more challenging for MARL algorithms while simultaneously improving the transferability of MARL models across different scenarios.

\subsection{LLM Script}

While MARL-based models have demonstrated impressive performance in competitive games, they frequently overfit to particular opponent strategies, leading to instability when encountering unfamiliar adversaries. In comparison, decision trees exhibit greater stability across diverse opponents and offer enhanced interpretability. Nonetheless, decision trees demand substantial prior knowledge and are susceptible to errors in edge cases, resulting in unforeseen outcomes.

\begin{figure}[h!]
    \centering
    \includegraphics[width=0.9\textwidth]{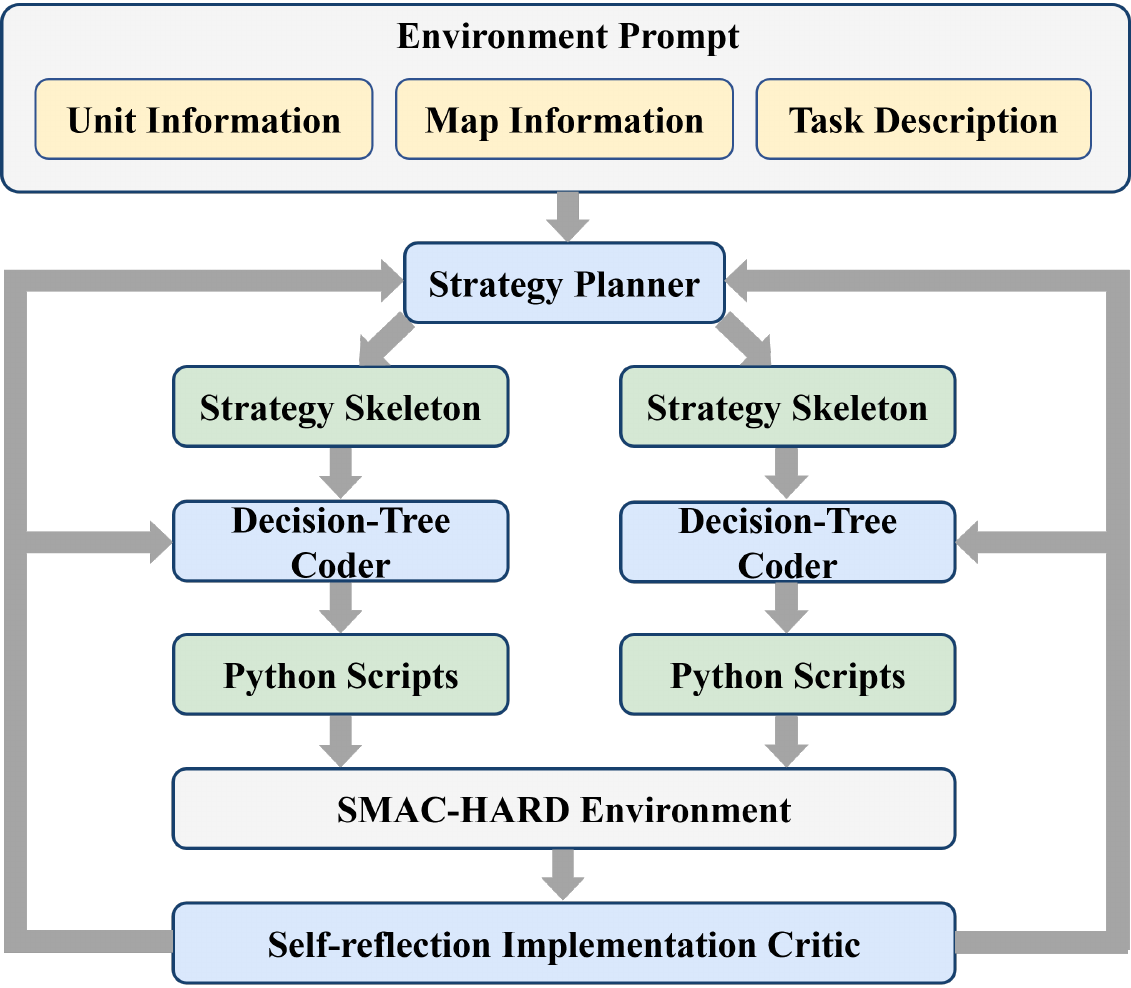}
    \caption{The unit information, map information, and task description serve as a system prompt and are passed to the planner. The planner plans strategy for both sides and the coders implement the strategy correspondingly. Then the python scripts are the red and blue side of SMAC-HARD to simulate. Finally, the critic module analyse the simulation results and provide refinement suggestions to the planner and the coders.}
    \label{fig:llm_arch} 
\end{figure}

Motivated by the recent work, LLM-SMAC \citep{deng2024new}, which leverages LLMs to address SMAC tasks, we adopt a similar pipeline for generating decision trees for both agents and opponents. As illustrated in Figure \ref{fig:llm_arch}, the agent and opponent share identical environment settings, encompassing map details, unit data, and task specifications. This shared information constitutes the environment prompt, which is fed into the strategy planner for both entities. Subsequently, the decision tree coder generates Python scripts for each side. Unlike LLM-SMAC, where the scripts are tailored for the python\_sc2 package, we convert the python\_sc2 scripts into pysc2 scripts using the deepseek coder model. Once the win rates for both sides stabilize, we select the script as the opponent script for our SMAC-HARD setup. The critic module evaluates the rollout outcomes and offers implementation suggestions and strategic refinements for both agents.

\subsection{Implementation}

\begin{figure}[h!]
    \centering
    \includegraphics[width=0.95\textwidth]{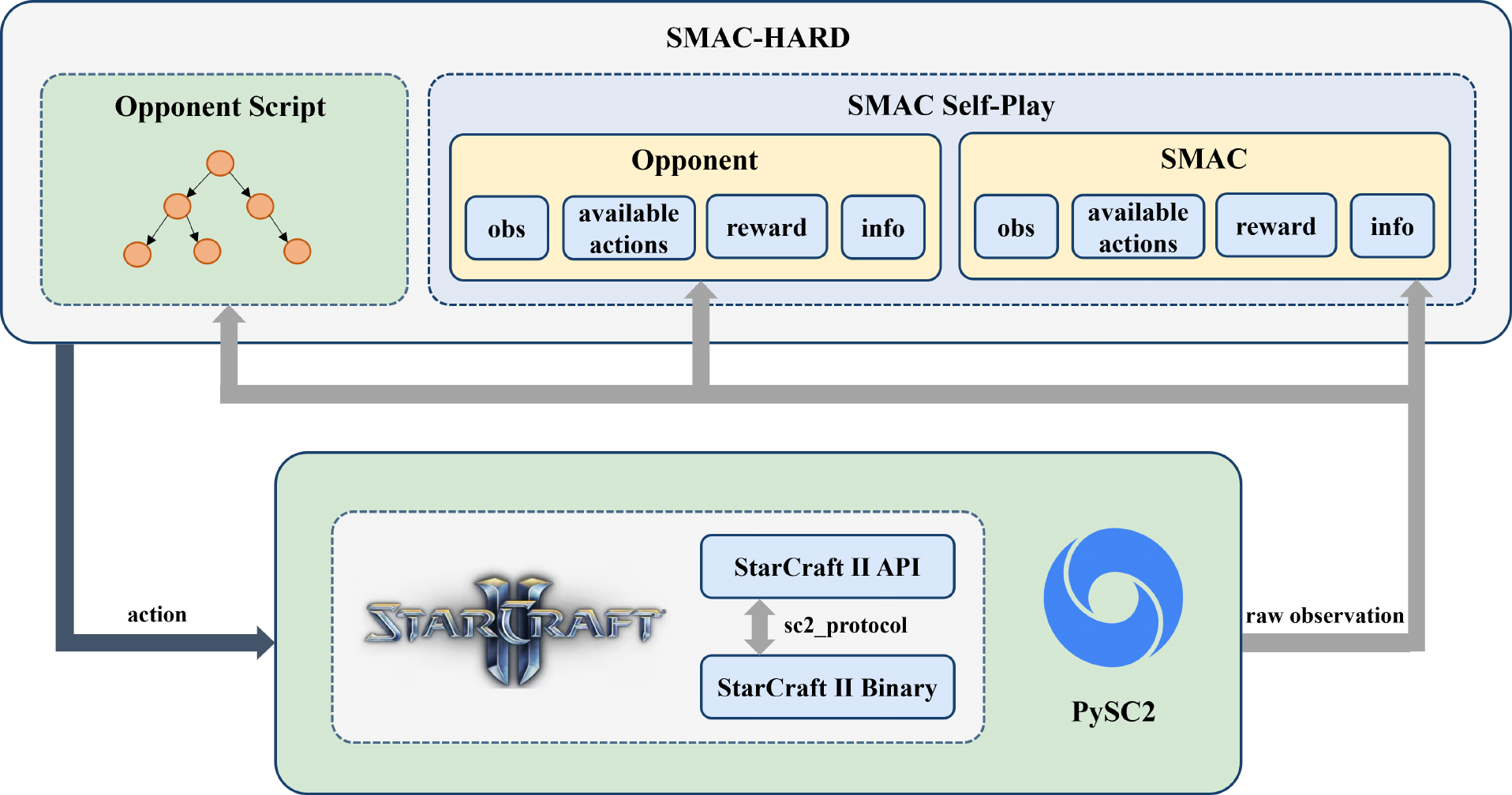}
    \caption{The overall architecture of our proposed SMAC-HARD, opponent decision script, self-play interface, and the SMAC, PySC2, StarCraft II modules.}
    \label{fig:hard-arch} 
\end{figure}

According to Figure \ref{fig:hard-arch}, in terms of source code, the pysc2 package serves as an abstraction of the sc2\_protocol, which is included in the StarCraft II binary files. Through pysc2, players can initiate new games, select maps, control units, and establish bases. The SMAC framework further encapsulates pysc2 by transforming the raw observations into structured, vectorized observations and states. Actions executed in SMAC are translated into commands that adhere to the sc2\_protocol API via pysc2. Consequently, the StarCraft II environment inherently supports both standardized actions from SMAC and actions generated by pysc2 scripts, provided they are converted into the sc2\_protocol API format.

Leveraging this inherent support, we modify the SC2Map to enable multi-player mode, ensuring that units controlled by the agent spawn at the Team 1 location for player 1, while opponent units spawn at Team 2 for player 2. To prevent action interference, we disable the default attack policy. Additionally, we reimplement the starcraft.py module in SMAC to accommodate two players, retrieve raw observations for both, and process actions from both players simultaneously. To mitigate the impact of action execution order, we parallelize the action-stepping process for the two players.

Beyond modeling the opponent’s decision tree, we introduce a random strategy selection function governed by predefined probability settings. These probabilities are represented as a list of float values, with equal probabilities set as the default configuration. Furthermore, we replicate the encapsulation of observations, states, and available actions for the agent and expose a similar interface for the opponent to facilitate self-play models. The self-play mode or decision-tree mode is controlled via a parameter, 'mode', which defaults to the decision-tree mode.

Under these conditions, users can seamlessly transition their experimental environment from SMAC to SMAC-HARD by simply updating the import command to 'import smac-hard'.

\section{SMAC-HARD Experiments}

In this section, we present experimental results for our proposed SMAC-HARD framework. Initially, we train widely used and state-of-the-art SMAC baselines on SMAC-HARD to evaluate the impact of the mixed opponent strategies. We select both value-based algorithms, such as QMIX, QPLEX, and LDSA, as well as policy-based algorithms like MAPPO and HAPPO, as our baseline methods. Additionally, we conduct black-box testing of these algorithms and provide detailed analysis. The final performance metrics and learning curves are included in this section, with supplementary details available in the Appendix.

\subsection{Baseline Comparisons}

\paragraph{Baseline} We select widely adopted and state-of-the-art algorithms for our experiments, including value-based methods QMIX and QPLEX, policy-based method MAPPO, and the latest actor-critic algorithm, HAPPO, all configured with their officially provided default parameters. The implementations of QMIX and QPLEX are sourced from the pymarl2 codebase \citep{hu2021rethinking}, while HAPPO and MAPPO are provided by \citep{yu2022surprising}. The LDSA is obtained from its respective codebases \citep{Yang_Zhao_Hu_Zhou_Li_2022}

\begin{figure}[h!]
    \centering
    \includegraphics[width=0.99\textwidth]{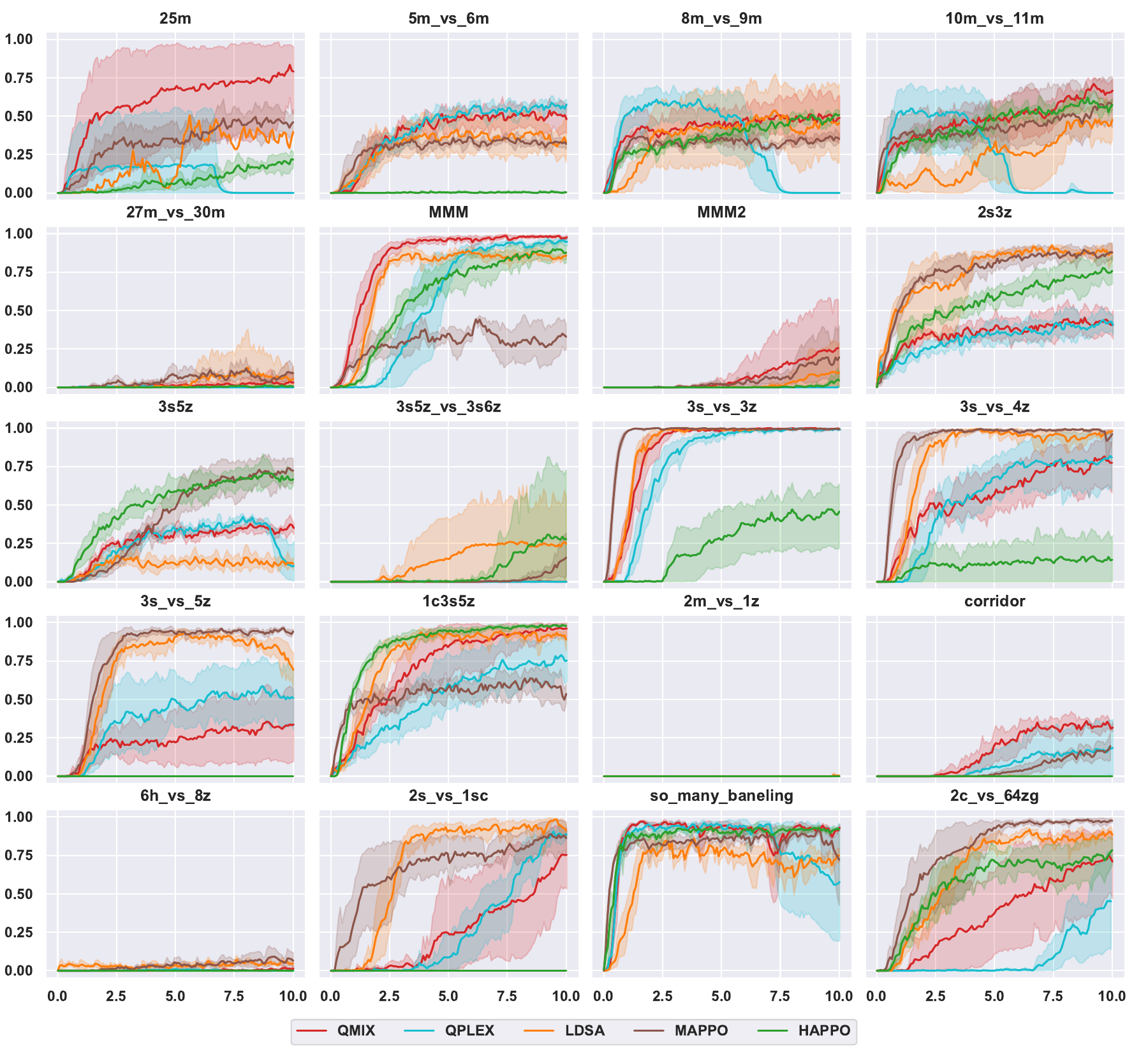}
    \caption{Comparison of mean test winning rate of different algorithms on new SMAC-HARD environments. The x-axis is the time steps (1e6) being evaluated and the y-axis is the average winning rate of 5 different seeds from 32 evaluation processes.}
    \label{fig:main-exp} 
\end{figure}

\paragraph{SMAC and SMAC-HARD} We evaluate the baseline algorithms on both the original SMAC environment and our proposed SMAC-HARD framework across 20 selected tasks. The difficulty level is set to 7 by default. The battle win rates are computed as the average of 32 evaluation runs. Each experiment is repeated five times with different random seeds, and the results are smoothed with a factor of 0.6 for improved visualization over 2 million and 10 million time steps, respectively. The shaded regions represent the variance across the five seeds, indicating the stability of the generated policies. The outcomes are presented in Figure \ref{fig:main-exp} and Table \ref{tab:hard-10m}.

\begin{table}[h!]

	\caption{Performance in SMAC-HARD tasks within 10M time steps}
	\label{tab:hard-10m}
	\centering

	\begin{tabularx}{\textwidth}{>{\centering\arraybackslash}p{3cm}
	>{\centering\arraybackslash}X
	>{\centering\arraybackslash}X
	>{\centering\arraybackslash}X
	>{\centering\arraybackslash}X
	>{\centering\arraybackslash}X}		

		\toprule
    	SMAC-HARD & QMIX & QPLEX & LDSA & MAPPO & HAPPO \\
    	\toprule
    	3m & 0.9938 & 0.0438 & 1 & 1 & 0.4375 \\
    	8m & 0.9628 & 0.0875 & 0.9687 & 0.6438 & 0.8625 \\
    	5m\_vs\_6m & 0.4375 & 0.6188 & 0.4 & 0.3138 & 0.004 \\
    	8m\_vs\_9m & 0.4938 & 0 & 0.7187 & 0.3528 & 0.5006 \\
    	10m\_vs\_11m & 0.6813 & 0 & 0.625 & 0.5681 & 0.5672 \\
    	25m & 0.7813 & 0 & 0.4375 & 0.4562 & 0.2092 \\
    	27m\_vs\_30m & 0.0313 & 0 & 0.125 & 0.0938 & 0.0063 \\
    	2s3z & 0.4125 & 0.4313 & 0.7737 & 0.8735 & 0.7547 \\
    	3s5z & 0.3625 & 0.1063 & 0.1876 & 0.7225 & 0.6691 \\
    	3s5z\_vs\_3s6z & 0 & 0 & 0.1319 & 0.1569 & 0.2573 \\
    	1c3s5z & 0.9625 & 0.7563 & 0.9375 & 0.5445 & 0.975 \\
    	3s\_vs\_3z & 0.9913 & 0.9875 & 1 & 0.9875 & 0.4875 \\
    	3s\_vs\_4z & 0.7938 & 0.7938 & 0.7815 & 0.9875 & 0.15 \\
    	3s\_vs\_5z & 0.3438 & 0.5 & 0.875 & 0.9407 & 0 \\
    	bane\_vs\_bane & 0.975 & 0.2313 & 0.8229 & 0.9125 & 0 \\
    	so\_many\_baneling & 0.9625 & 0.5938 & 0.875 & 0.8813 & 0.9188 \\
    	2s\_vs\_1sc & 0.7563 & 0.8875 & 0.4375 & 0.8375 & 0 \\
    	2m\_vs\_1z & 0 & 0 & 0 & 0 & 0 \\
    	2c\_vs\_64zg & 0.7891 & 0.5 & 0.9062 & 0.9741 & 0.7686 \\
    	MMM & 0.9875 & 0.95 & 0.875 & 0.3267 & 0.8813 \\
    	MMM2 & 0.275 & 0.0006 & 0.5312 & 0.1925 & 0.0425 \\
    	6h\_vs\_8z & 0.0188 & 0 & 0.0625 & 0.0607 & 0 \\
    	corridor & 0.3063 & 0.1688 & 0 & 0.1804 & 0 \\
    	\bottomrule
	\end{tabularx}

\end{table}

According to the results presented in Figure \ref{fig:main-exp}, the mixed opponent strategies significantly increase the difficulty for current MARL algorithms. In previously straightforward tasks, such as 2s3z, 3s5z, and 2s\_vs\_1sc, the convergence speed drops by up to $100\%$. In more challenging tasks, the mixed opponent strategies shift the convergence points among MARL algorithms, highlighting differences in their optimality. In super-hard scenarios, including MMM2, 3s5z\_vs\_3s6z, 27m\_vs\_30m, and corridor, not all baseline MARL algorithms achieve satisfactory final performance. These observations suggest that MARL algorithms with broader policy coverage capabilities are required to address these challenges effectively.

In comparison to the original SMAC tasks, where nearly all algorithms achieve close to $100\%$ win rates within 10 million time steps, SMAC-HARD introduces significantly higher difficulty, as evidenced by the lower win rates. As shown in Figure \ref{fig:main-exp}, the 2m\_vs\_1z task, which is relatively easy in the original SMAC environment, becomes a super-hard task in SMAC-HARD. We analyzed this phenomenon by reviewing replay videos. In SMACv1, the opponent zealot's attack target changes based on the hate values of the two marine agents, causing the zealot to focus on the marine currently attacking. This leads to a strategy where the two marines alternate attacks to divert the zealot's attention. However, in SMAC-HARD, the zealot consistently targets one marine, requiring one marine to kite the enemy while the other focuses on attacking. This dominant strategy demands precise and continuous action from the targeted marine, posing a significant challenge to MARL algorithms.

In addition to the conservative Q-values introduced by the mixed opponent strategies, the rollout returns exhibit higher variance. This increased variance poses challenges for MARL algorithms, such as the collapse of attention matrices after convergence during training, which may affect attention-based MARL algorithms. To illustrate this phenomenon, we provide the win rates of baseline algorithms at 2 million time steps in the Appendix, which could serve as a potential benchmark for evaluating the sample efficiency of new MARL algorithms.

\subsection{Black-box Evaluation}

To assess the transferability of models trained using MARL algorithms, we conducted a black-box evaluation. During training, agents were exposed only to the default opponent strategy on the original maps, but during evaluation, they faced the new mixed opponent strategies in the SMAC-HARD environment. We trained the baseline algorithms on the SMACv1 environment for 10 million time steps and saved the trained models. These models were then tested in the SMAC-HARD environment to measure their win rates. The black-box evaluation results are summarized in Table \ref{tab:black}.

\begin{table}[h!]

	\caption{Performance in SMAC-HARD tasks in black-box mode.}
	\label{tab:black}
	\centering

	\begin{tabularx}{\textwidth}{>{\centering\arraybackslash}p{3cm}
	>{\centering\arraybackslash}X
	>{\centering\arraybackslash}X
	>{\centering\arraybackslash}X
	>{\centering\arraybackslash}X}		

		\toprule
    	SMAC-HARD & QMIX & QPLEX & MAPPO & HAPPO \\
            \toprule
            3m & 0.0117 & 0.0898 & 0.1028 & 0.0573 \\
            8m & 0.0375 & 0 & 0.0513 & 0.1217 \\
            5m\_vs\_6m & 0 & 0 & 0 & 0 \\
            8m\_vs\_9m & 0 & 0 & 0 & 0 \\
            10m\_vs\_11m & 0 & 0 & 0 & 0 \\
            25m & 0 & 0 & 0 & 0 \\ 
            27m\_vs\_30m & 0 & 0 & 0 & 0 \\
            2s3z & 0.0508 & 0 & 0.1217 & 0.0959 \\
            3s5z & 0.0313 & 0 & 0.0375 & 0.0705 \\
            3s5z\_vs\_3s6z & 0 & 0 & 0 & 0 \\ 
            1c3s5z & 0.0703 & 0 & 0.1027 & 0.0912 \\
            3s\_vs\_3z & 0.3359 & 0.3555 & 0.4125 & 0.3724 \\
            3s\_vs\_4z & 0.7461 & 0.5052 & 0.4467 & 0 \\
            3s\_vs\_5z & 0.5305 & 0.8573 & 0.5359 & 0 \\
            bane\_vs\_bane & 0.2578 & 0 & 0.3484 & 0 \\
            so\_many\_baneling & 0 & 0 & 0 & 0 \\
            2s\_vs\_1sc & 0 & 0 & 0 & 0 \\
            2m\_vs\_1z & 0 & 0 & 0 & 0 \\
            2c\_vs\_64zg & 0.6238 & 0 & 0.3419 & 0.2565 \\
            MMM & 0 & 0 & 0 & 0 \\
            MMM2 & 0 & 0 & 0 & 0 \\
            6h\_vs\_8z & 0.0898 & 0 & 0 & 0 \\
            corridor & 0 & 0 & 0 & 0 \\
    	\bottomrule
	\end{tabularx}

\end{table}

The black-box evaluation reveals that MARL algorithms tend to overfit to the specific single opponent strategy they encounter during training, as evidenced by the low evaluation win rates. This suggests that the skills learned by these models are not robust or generalizable strategies but rather aggressive or opportunistic tactics tailored to the specific training scenario.

An illustrative example can be found in the comparison of win rates across the 3s\_vs\_3z, 3s\_vs\_4z, and 3s\_vs\_5z tasks. Unlike other scenarios where MARL models achieve near-zero win rates, the black-box evaluation win rates increase with the difficulty of the tasks. To effectively solve the stalker versus zealot task, the optimal strategy involves kiting—stalker units must alternate between walking and attacking to exploit their speed advantage. In the easier 3s\_vs\_3z task in SMACv1, agents may learn aggressive attacking skills that are sufficient to win. However, for more challenging tasks, agents must adopt a strict kiting mechanism, which is the optimal response strategy. Consequently, agents that have learned the optimal response strategy are more likely to succeed against the black-box edited opponent scripts. This phenomenon underscores the critical importance of training with a diverse range of strategies to ensure robust and generalizable models.

\section{Conclusion}

In this study, we present a series of experimental evaluations to demonstrate that the single, default opponent policy used in SMAC and SMACv2 lacks diversity in policy spaces. To address this limitation, we introduce SMAC-HARD, which supports opponent script editing, probabilistic mixed opponent policies, and self-play interface alignment, significantly mitigating the issue. Our results show that even popular and state-of-the-art MARL algorithms, which achieve near-perfect performance in traditional SMAC environments, struggle to maintain high win rates in SMAC-HARD. Additionally, we conduct a black-box evaluation of models trained using MARL algorithms to highlight the limited transferability of strategies when facing a single, vulnerable opponent policy. Finally, we align the training interface for opponents with that of the agents, providing a platform for potential self-play research in MARL. We believe that SMAC-HARD can serve as a challenging and editable domain, contributing to the MARL research community by capturing practical challenges and fostering further advancements.

\newpage
\bibliography{neurips_2024}
\bibliographystyle{neurips_2024}

\newpage
\appendix
\section*{Appendix}
\section{Final Performance at 2M time step}

In the Baseline Comparisons section above, we have listed the final performance of baseline algorithms at 10M time steps. Additionally, we also list the performance at 2M time step. The results may also considered as the benchmark to judge the sample efficiency of an algorithm.

\begin{table}[!ht]
    \small
    \centering
    \caption{Final performance at 2M time step with \textbf{default opponent policy}}
    \begin{tabularx}{\textwidth}{>{\centering\arraybackslash}p{3cm}
	>{\centering\arraybackslash}X
	>{\centering\arraybackslash}X
	>{\centering\arraybackslash}X
        >{\centering\arraybackslash}X
	>{\centering\arraybackslash}X}
        \toprule
        SMAC & QMIX & QPLEX & LDSA & MAPPO & HAPPO \\
        \toprule
        3m & 0.9798 & 0.9856 & 1 & 0.9897 & 0.9945 \\
        8m & 0.9797 & 0.972 & 0.9684 & 0.9538 & 0.9947 \\
        5m\_vs\_6m & 0.544 & 0.4385 & 0.7605 & 0.5103 & 0.437 \\
        8m\_vs\_9m & 0.9137 & 0.6479 & 0.9308 & 0.7821 & 0.3976 \\
        10m\_vs\_11m & 0.9515 & 0.6542 & 0.929 & 0.6859 & 0.2822 \\
        25m & 0.975 & 0.5292 & 0.9655 & 0.9692 & 0.94 \\
        27m\_vs\_30m & 0.4202 & 0.1245 & 0.5161 & 0.6051 & 0.8289 \\
        2s3z & 0.9732 & 0.9796 & 0.9843 & 0.941 & 0.9831 \\
        3s5z & 0.9485 & 0.9351 & 0.9301 & 0.4512 & 0.9695 \\
        3s5z\_vs\_3s6z & 0.0256 & 0.0783 & 0.5468 & 0.1128 & 0.0738 \\
        1c3s5z & 0.9893 & 0.9607 & 0.9834 & 0.982 & 0.9771 \\
        3s\_vs\_3z & 0.994 & 0.9945 & 0.9991 & 0.982 & 0.9846 \\
        3s\_vs\_4z & 0.9857 & 0.364 & 0.9997 & 0.9744 & 0.6434 \\
        3s\_vs\_5z & 0.7771 & 0.3729 & 0.9137 & 0.9744 & 0.2051 \\
        bane\_vs\_bane & 0.783 & 0.9967 & 1 & 0.9974 & 0.9862 \\
        so\_many\_baneling & 0.9672 & 0.9504 & 0.9685 & 0.9821 & 0.9792 \\
        2s\_vs\_1sc & 0.9917 & 0.9906 & 0.9943 & 1 & 0.9965 \\
        2m\_vs\_1z & 0.9926 & 0.988 & 1 & 1 & 0.9936 \\
        2c\_vs\_64zg & 0.9226 & 8318 & 0.9425 & 0.9359 & 0.9494 \\
        MMM & 0.9824 & 0.9783 & 0.9743 & 0.9256 & 0.989 \\
        MMM2 & 0.7875 & 0.2747 & 0.7582 & 0.4487 & 0.9068 \\
        6h\_vs\_8z & 0.1438 & 0.0074 & 0.1362 & 0.0154 & 0 \\
        corridor & 0 & 0 & 0.8234 & 0.3077 & 0.3693 \\
        \bottomrule
    \end{tabularx}
\end{table}

\begin{table}[!ht]
    \small
    \centering
    \caption{Final performance at 2M time step with \textbf{mixed edited opponent policy}}
    \begin{tabularx}{\textwidth}{>{\centering\arraybackslash}p{3cm}
	>{\centering\arraybackslash}X
	>{\centering\arraybackslash}X
	>{\centering\arraybackslash}X
        >{\centering\arraybackslash}X
	>{\centering\arraybackslash}X}
        \toprule
        SMAC & QMIX & QPLEX & LDSA & MAPPO & HAPPO \\
        \toprule
        3m & 0.9917 & 0.9958 & 1 & 0.6166 & 0.35 \\
        8m & 0.8083 & 0.8792 & 0.4375 & 0.5979 & 0.7687 \\
        5m\_vs\_6m & 0.3166 & 0.2688 & 0.2563 & 0.3104 & 0 \\
        8m\_vs\_9m & 0.4063 & 0.5729 & 0.3646 & 0.2958 & 0.2938 \\
        10m\_vs\_11m & 0.3646 & 0.5292 & 0.1632 & 0.3813 & 0.35 \\
        25m & 0.5313 & 0.1563 & 0.1563 & 0.2583 & 0.0063 \\
        27m\_vs\_30m & 0 & 0.0042 & 0.0083 & 0.0125 & 0 \\
        2s3z & 0.3438 & 0.2688 & 0.6542 & 0.7042 & 0.4229 \\
        3s5z & 0.2312 & 0.1292 & 0.1146 & 0.0667 & 0.3417 \\
        3s5z\_vs\_3s6z & 0 & 0.0021 & 0 & 0 & 0 \\
        1c3s5z & 0.4708 & 0.2729 & 0.7042 & 0.5292 & 0.852 \\
        3s\_vs\_3z & 0.8771 & 0.5458 & 0.9497 & 0.9979 & 0.0021 \\
        3s\_vs\_4z & 0.4437 & 0.1938 & 0.8247 & 0.95 & 0.1063 \\
        3s\_vs\_5z & 0.2125 & 0.3021 & 0.566 & 0.7479 & 0 \\
        bane\_vs\_bane & 0.9646 & 0.9667 & 0.8125 & 0.8813 & 0.9938 \\
        so\_many\_baneling & 0.9375 & 0.9458 & 0.7313 & 0.7896 & 0.8729 \\
        2s\_vs\_1sc & 0.0167 & 0.0104 & 0.2896 & 0.5604 & 0 \\
        2m\_vs\_1z & 0 & 0 & 0 & 0 & 0 \\
        2c\_vs\_64zg & 0.0958 & 0.0021 & 0.3333 & 0.6542 & 0.4417 \\
        MMM & 0.8167 & 0.0167 & 0.7153 & 0.25 & 0.2854 \\
        MMM2 & 0 & 0 & 0 & 0 & 0 \\
        6h\_vs\_8z & 0 & 0 & 0.023 & 0.0125 & 0 \\
        corridor & 0 & 0 & 0 & 0 & 0 \\
        \bottomrule
    \end{tabularx}
\end{table}

\section{Return Performance}

\begin{figure}[h!]
    \centering
    \includegraphics[width=0.99\textwidth]{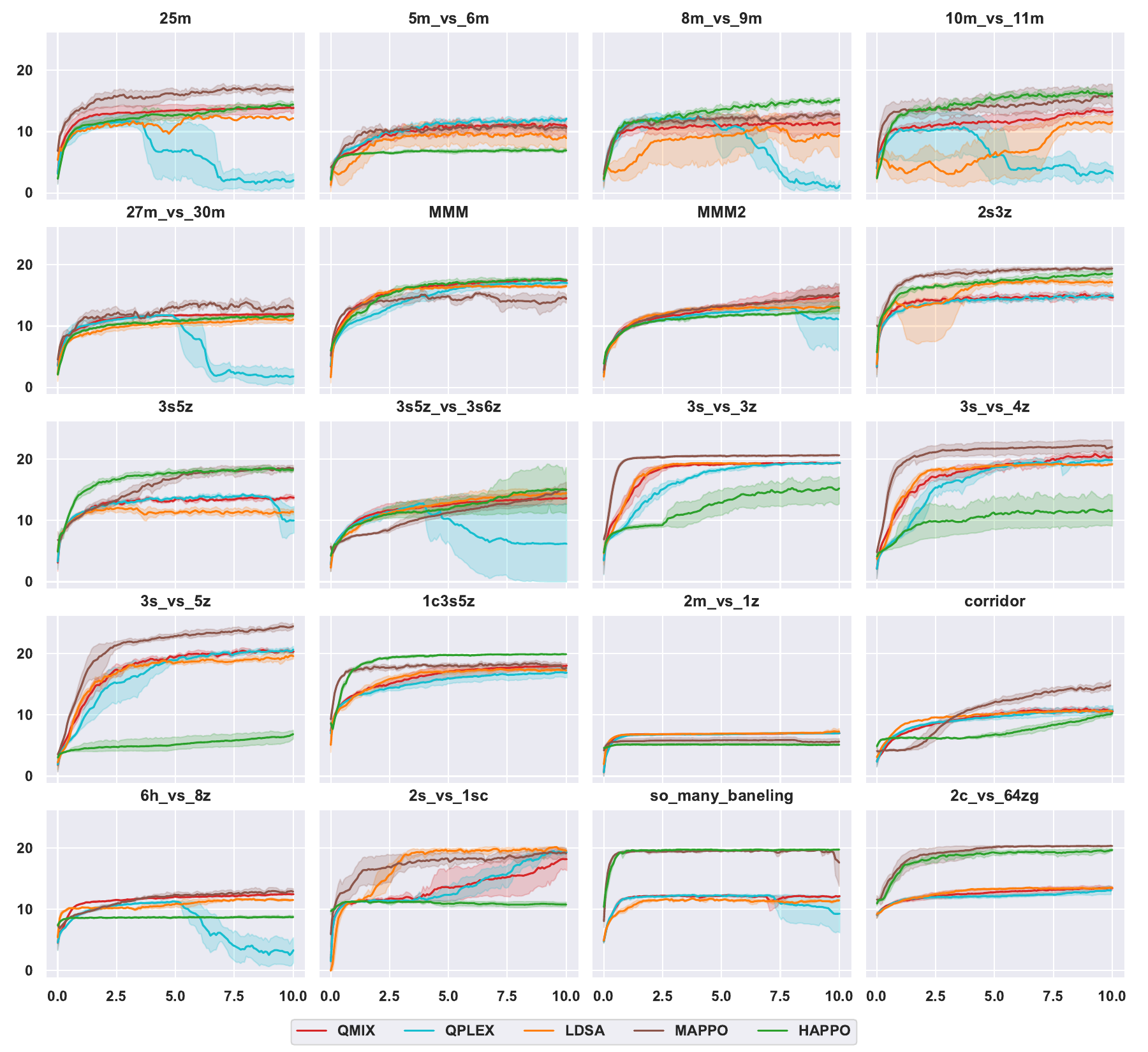}
    \caption{Comparison of mean test expected return of different algorithms on new SMAC-HARD environments. The x-axis is the time steps (1e6) being evaluated and the y-axis is the average winning rate of 5 different seeds from 32 evaluation processes.}
    \label{fig:return} 
\end{figure}

In line with the reward calculation methods used in SMAC, there is a general trend where higher returns correlate with higher win rates. However, the reward is determined by multiple factors, including health and shield values, while the win condition is based on which side has no remaining units. This discrepancy can lead to a slight misalignment between the expected return and the actual win rate. For instance, a strategy that results in a single full-health agent surviving would yield significantly higher rewards compared to one where three agents survive but with low health. In this context, we present the expected returns in this section as shown in Figure \ref{fig:return}.

\end{document}